# Applying graph matching techniques to enhance reuse of plant design information

## Abstract

This article investigates how graph matching can be applied to process plant design data in order to support the reuse of previous designs. A literature review of existing graph matching algorithms is performed, and a group of algorithms is chosen for further testing. A use case from early phase plant design is presented. A methodology for addressing the use case is proposed, including graph simplification algorithms and node similarity measures, so that existing graph matching algorithms can be applied in the process plant domain. The proposed methodology is evaluated empirically on an industrial case consisting of design data from several pulp and paper plants.

## 1 Introduction

Engineering design is a highly competitive field where the ability to develop high-quality, low-cost results in less time is needed to gain advantage in the market. Reusing knowledge gained from previous projects and designs can be helpful in this task [1-2]. Despite providing many benefits during the design process, reusing information also has its challenges. One of the primary concerns is the retrieval of relevant information [3]. Even if useful information from previous projects is available, it holds no value for the designers if they cannot easily find and apply it [4]. Discovery of relevant information is therefore an essential part of the design reuse process.

The following use case was provided by our industrial partner, which also provided the research materials for this article. In plant engineering, the parametrization of the process components takes place early in the design process of a new plant. At this stage, the structure of some process area has already been designed: the main components and the connections between them are known. This information is typically represented in the form of a piping and instrumentation diagram (P&ID). However, the level of detail is still rough: for example, the pump and valve types, the dimensions of the components, and the process parameters, such as temperature and pressure, still need to be selected. This information is crucial, as the design process of the plant consists of multiple subprocesses working in parallel. Some of these subprocesses, for example initial cost and material estimations, require information about the said parameters before they can continue forward. This forces the designers to make approximate estimations for the parameter values, so that the parametrization phase does not create a bottleneck in the project. The timely discovery of estimates for the parameters thus has a significant impact on the entire project schedule, and the quality of the estimates impacts the work of the other design processes.

As the intention is to find parameters for the components, they do not yet have many attributes that could aid with the identification. For this reason, the main source of information lies in the structure of the process: what components are included and how they are connected to each other. Graph-based methods offer a way for finding these correspondences, since process plants have a natural graph representation: process components, such as tanks, pumps and pipe segments can be seen as nodes of a graph, in which the flows from one process component to another are represented by directed edges.

The research goal of this article is to propose a method of finding automatically a process component from a previous project that is similar to a component in the current one applying graph matching methods. Similar methods have been used in various domains, such as image recognition [5], social networks [6] and biology [7]. However, the use of graph matching with process plants is a less researched area. [8] matches CAD models, so the focus is on the geometry of components rather than the process. [9-10] model the process as a graph in a similar way as this article. However, the focus of both papers is not on finding component level correspondences. Thus, the application of graph matching methods to our use case is an unsolved problem. Further, the following challenges need to be addressed:

1. The matched sites contain design differences and the source data can contain errors, which means that finding an exact match between the graphs is not possible.
2. The limited amount of information available in the beginning of the project presents another difficulty.
3. The large size of the graphs poses a scalability challenge
4. Graph matching algorithms designed for large graphs generally do not take edge labels and directions into account, but for process plant applications this information is needed to capture flow type and direction

# 2 Related research

## 2.1 Graph matching

Graph matching aims to find corresponding nodes between two or more graphs [11-14]. Exact graph matching finds an edge-preserving mapping between the nodes of the matched graphs. This approach is not applicable when the graphs are known to be different, so the goal is to find similarities rather than perfect matches [15]. This is also the case with our goal of finding reusable engineering data from one-of-a-kind plant delivery projects. Inexact graph matching algorithms can handle differences between the matched graphs [11-14]. Two common approaches for inexact graph matching include the quadratic assignment problem (QAP) [16-18] and the graph edit distance (GED) [19-20]. Both are NP-hard, but the computational complexity can be reduced with heuristics, in which case the obtained match is not guaranteed to be optimal [21-23]. For our research problem, a heuristic is necessary due to scalability reasons, since the graphs have up to thousands of nodes. A non-optimal solution is acceptable, since the intention is not to automate the expert human designer, but to provide tools to reduce the number of engineering hours spent searching for reusable engineering data.

In some circumstances, the requirement to find one-to-one matches between the nodes of the two graphs is too restrictive. For example, one pulp-and-paper plant may have two refiners to perform a refinement function while another plant may have three. In such cases, a many-to-many matching approach may give better performance [24].

Another variation of matching algorithms is Local Network Alignment (LNA), which searches for similar substructures between the graphs and often produces many-to-many matchings [25]. The found regions are often small in size and some areas can remain unmatched [26]. However, our initial experiments on applying LNA to our problem produced unsatisfactory results, so LNA was not investigated further.

Pair-wise algorithms search for matches between two graphs, whereas multi-graph algorithms try to find a consistent mapping across a set of graphs. Many of these algorithms act more like frameworks that can incorporate pair-wise matching algorithms, such as [27-30]. In our research material, only two examples of each process plant type were available, so only pair-wise graph matching algorithms were used. Multi-graph matching remains a topic for further research.

Anchors, also known as seeds, are corresponding node pairs that have been identified from the graphs before the matching begins. Anchors can be found, for example, by a human expert [31]. In some algorithms, even a small number of anchor nodes can be used to aid the graph matching process [31-33].

The performance of many graph matching algorithms can be improved by a similarity measure that quantifies the similarity between two nodes or edges from different graphs. The way that the similarity values are calculated depends on the domain. In protein interaction networks, BLAST E-values that describe the sequence similarity between two proteins are used [15]. String comparison is used between the names of tables and columns of relational schemas [34]. In a UML class diagram matching application [35], the similarity between two classes was calculated by comparing the class names and the number of attributes and methods. In image processing, the Euclidean distance between the points has been used [36]. None of these is applicable to our domain, so a new similarity measure is proposed in this article.

Whereas similarity measures look at a pair of nodes or edges, alignment quality measures quantify the goodness of the mapping between two graphs. Some graph matching algorithms are able to use them as optimization measures. The most common used ones are Edge Correctness (EC) [37], Induced Conserved Structure (ICS) [38], Symmetric Substructure Score S$^3$ [39] and Weighted Edge Correctness (WEC) [40].

Table 1 presents an overview of relevant algorithms from different domains. Only algorithms that scale to graphs consisting of hundreds of nodes have been included. The ability to incorporate node attribute information, edge labels or edge direction was considered desirable. As the much researched protein-protein interaction (PPI) network algorithms fill many of these requirements, they are the most represented group. Image matching is another popular field in graph matching, but many of the commonly used algorithms require a large compatibility matrix as an input [41-43], limiting scalability. As noted in [44], these algorithms have been tested with graphs of 20-200 nodes, significantly smaller than the graphs under consideration in our use case.

*Table 1 Overview of the graph matching algorithms*

| Name | Group | Original application domain | Selected for empirical testing |
|---|---|---|---|
| DSPFP [44] | QAP | Images | x |
| PATH [24] | QAP | Images | |

| NATALIE 2.0 [45] | QAP | PPI | |
|---|---|---|---|
| L-GRAAL [46] | QAP | PPI | |
| Similarity flooding [34] | Iterative similarity | Schemas | x |
| Blondel et al. [47] | Iterative similarity | Synonyms | |
| Zager and Zerghese [48] | Iterative similarity | Synthetic | |
| IsoRank [49] | Iterative similarity | PPI | |
| WAVE [40] | Seed-and-extend | PPI | x |
| GHOST [38] | Seed-and-extend | PPI | |
| GRAAL [37] | Seed-and-extend | PPI | |
| MI-GRAAL [50] | Seed-and-extend | PPI | |
| TALE [51] | Seed-and-extend | PPI | |
| HubAlign [52] | Seed-and-extend | PPI | |
| ExpandWhenStuck [33] | Seed-and-extend | Social networks | |
| GEDEVO [53] | GED | PPI | x |
| Riesen and Bunke [54] | GED | Various | |
| Bougleux et al. [55] | GED | Chemistry | |
| SANA [56] | Other | PPI | x |
| MAGNA++ [57] | Other | PPI | |
| Optnetalign [58] | Other | PPI | |
| NETAL [59] | Other | PPI | |
| PISwap [60] | Other | PPI | |
| ModuleAlign [52] | Other | PPI | |
| IGLOO [61] | Other | PPI | |

Five of the graph matching algorithms presented in Table 1 were chosen for empirical testing. Table 1 has a 'group' column and one algorithm from each group was selected. Further, algorithms from different application domains were selected. Additionally, the following three criteria were used to guide the selection. Firstly, all algorithms should allow the user to specify the node similarity values in the range [0,1]. Secondly, they should be capable of handling graphs with 1000 nodes. Thirdly, either the authors should have provided an implementation of the algorithm online, or the original article should contain enough instructions for self-implementing the algorithm.

## 2.2 Process systems engineering

Appropriate application of graph matching techniques requires the consideration of process systems engineering (PSE) [62] as an activity consisting of sequential or parallel tasks. The goals of a specific task need to be considered, so that graph matching methods can be proposed to meet those goals, which are more narrowly scoped than the goals of the entire process systems engineering activity. The availability of source information for that task is a constraint that must be respected. This decomposition of goals and introduction of constraints may result in designs that are suboptimal. However, not all articles on PSE give an explicit description of how the activity is decomposed to tasks, and the same decomposition may not apply to all PSE methodologies (e.g. [63,64]). In contrast, [65] explicitly discusses the possible design approaches in terms of a sequence of engineering activities, and reflects on the implications from the perspective of obtaining an optimal design.

With respect to the use case stated in section 1, it is important to emphasize that this article is not concerned with designing the process structure. Rather, the focus is on the initial parameter estimation phase, which occurs after an optimal or suboptimal process structure has been obtained. In our use case, it is assumed that process structure design and initial parameter estimation are distinct, consecutive phases. The industrial partner that provided the case is a global leader in the field and the assumption applies to their practice. Whether it applies to specific PSE research articles is difficult to ascertain, since these aspects are not focal to the argument. For example, [66] discusses a sequential state-of-the-art PSE methodology and [67] explicitly evaluates alternative process structures, but it is unclear whether the process structure design and the parameterization of the process components are jointly addressed or if they are separate activities producing possibly suboptimal results.

In this paper, the focus is specifically about accelerating PSE activities by automating the search of reusable previous design data in the initial parameter estimation task. Research on design information reuse in PSE is limited. PSE ontology reuse [68] is not relevant to the said focus. [69] explores the possibility of reusing components, but it is unclear if the designs are optimal from the PSE perspective, and in any case the approach is not applicable to initial parameter estimation. Despite the lack of research, it is clear that the reuse of previous designs may introduce constraints and thus result in suboptimal solutions. Nevertheless, according to our industrial partner, reuse of designs in PSE is common in the industry, since it can substantially shorten delivery times and lower the engineering cost, which is crucial from the perspective of the designer. Therefore, the approach presented in this paper is aimed at PSE practitioners already engaged in reusing previous designs. The contribution of the paper is not related to helping such practitioners avoid suboptimal solutions but to automate their search for relevant previous projects and thus reduce the number of engineering hours they spend on the initial parameter estimation task.

Finally, the relation of our proposed graph representation to PSE artifacts is as follows. Our methodology involves constructing a 'process graph' from the available source information of a site, after which our proposed graph matching methodology is applied to the graphs of two sites to be matched. Our 'process graph' introduced in section 3.2 has some differences to the P-graph commonly used in PSE [70]. The P-graph is not well suited to capturing the available machine-readable source information, nor does it permit effective ways of reducing the number of nodes, which is critical to our methodologies scalability to large, industrial scale processes. In particular, our process graph needs to accommodate a pipeline consisting of several pipes or elbows connected to each other without any equipment nodes in between, since key sources of information such as 3D CAD databases store the pipeline data in such a form. Further, some of our simplification algorithms may remove all the pipelines from the graph, in which case it would not be a P-graph.

# 3 Research material

Data was obtained from four plants referred to as sites A, B, C and D. The process structure of site A resembled site B, and site C resembled site D, so graph matching was applied between A and B and between C and D. The data was from completed or almost completed projects. To investigate the applicability of our method in the initial design stage, data was gathered from the early stages of site A's development, referred to as early site A.

Sites A and B were both multi-ply board plants that produced board with three layers. Both sites contained additional large systems not found in the other site. The scope of site A included a coating color kitchen not found in site B, and site B included a steam and condensate system not found in site A. This resulted in a high number of outliers, which are components that do not have a match in the other site. Sites C and D both consisted of a recycled fiber plant that was connected to a linerboard process.

While the compared sites followed the same general structure, they were not identical. For purposes of illustration, consider the schematic drawing of a subsection of a multi-board plant in Figure 1. This does not correspond to any specific customer plant, but rather has features commonly found in this kind of plant. In particular, there is a bottom, middle and top line for the three layers of the multi-board. The drawing gives the reader an idea of the similarities and differences between sites. Considering the numerous chests and refiners, their connectivity may differ between two sites: even if two components were connected in the first site, the components that they corresponded to in the second site were not necessarily connected. While the sites generally shared the same subsections, their internal structures had differences. The most common type of difference was in the number of components performing the same task, and whether the components were connected in series or parallel. This applies to many components, such as the towers, chests and refiners in Figure 1 as well as other kinds of components such as screens and filters. An example graph representation of differences is presented in Figure 2.

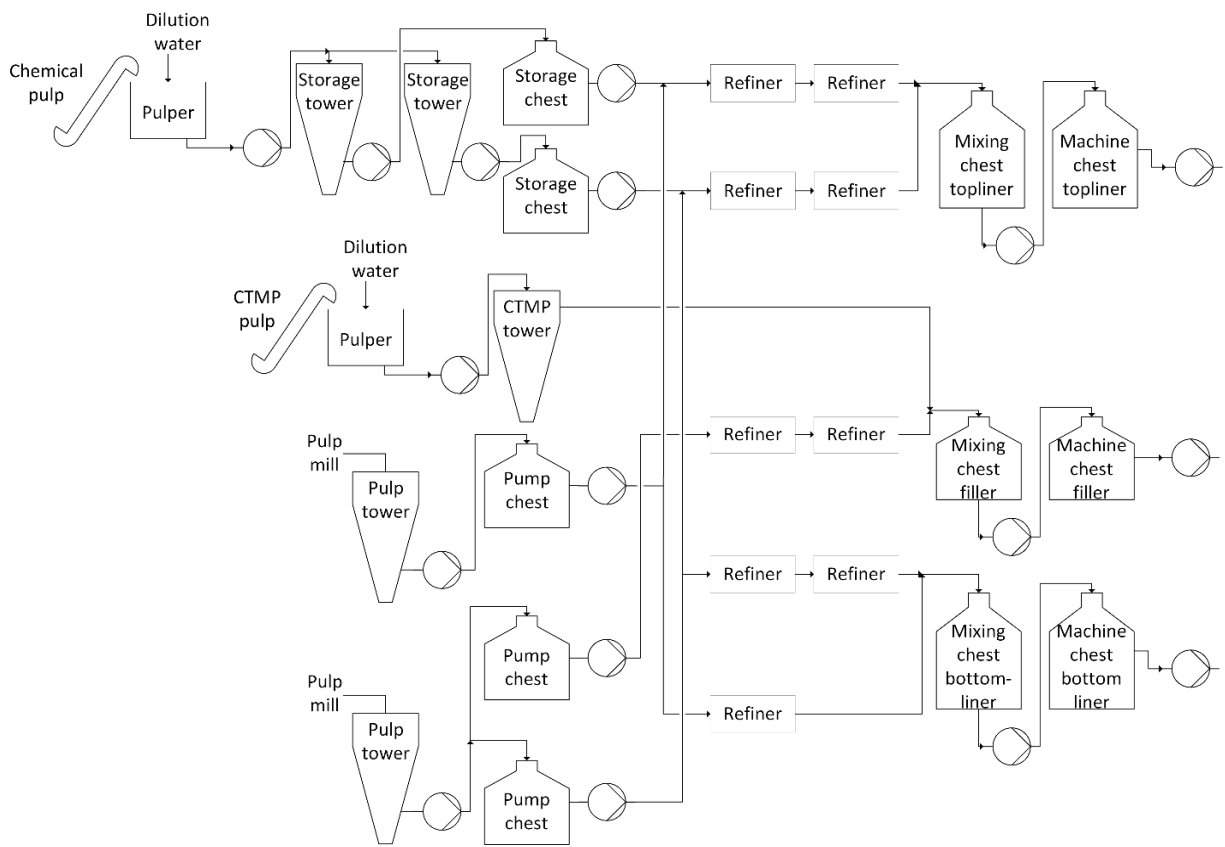

*Figure 1 Schematic drawing of a subsection of a multi-board plant*

**Site 1**

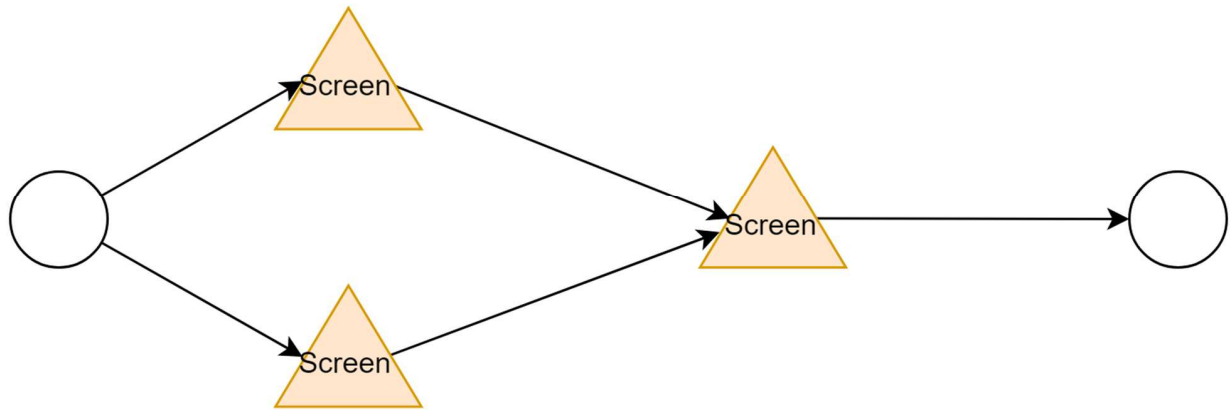

**Site 2**

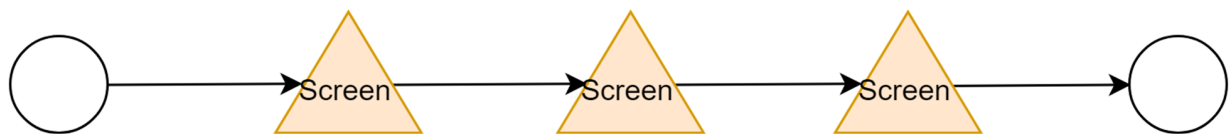

*Figure 2 Graph representation of an example of differences in corresponding structures between two sites. A varying number of components can be connected in series or in parallel.*

Due to the nature of our use case, the proposed methodology should be applicable as early in the plant design process as possible. The graphs for sites A and B were created from 3D plant design data. The graphs for sites C and D were created based on data gathered from human-readable line list data. The line list format was especially prone to errors such as duplicate nodes. For all graphs, data cleaning was performed before the matching. This mostly consisted of finding duplicate nodes and merging them as in Figure 3 – the figure shows a tank, but the same approach applied to other process components such as the chests, towers and refiners in Figure 1. However, as the graphs consisted of thousands of nodes and edges, only the most obvious errors were fixed.

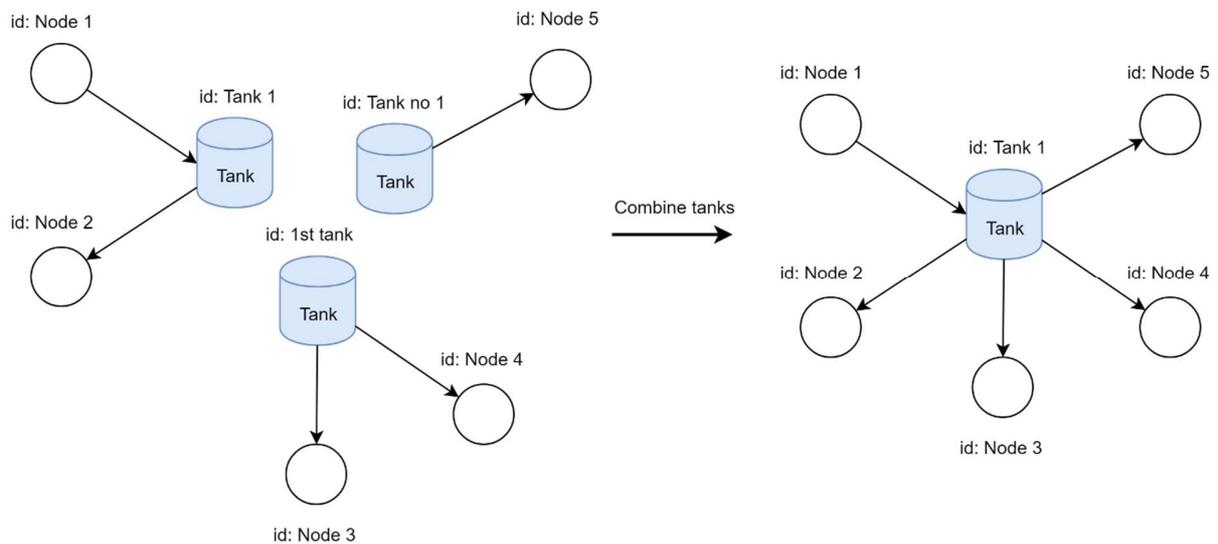

*Figure 3 Left side shows how spelling differences for the identifiers of a single component can create duplicate nodes. The right side shows a fixed graph where all the tank nodes are combined into one*

# 4 Proposed methodology

This section describes a method for matching process plant models. The method has the following steps:

1. Translate process plant models to process graphs. The implementation of this step depends on the original data format of the plant models.
2. Simplify the process graphs by removing some nodes while maintaining the overall topology.
3. Calculate node similarity measure for all pairs of source and target graph nodes.
4. Apply a general purpose graph matching algorithm to the simplified source and target graphs with the calculated node similarity measure.

## 4.1 Building the process graph

### 4.1.1 Process graph description

The process graph is the graph representation of a process plant model as a directed graph with node and edge labels. The nodes of the graph correspond to the components of the process plant. The type of a node can be unknown if it does not belong to the above list or if it could not be identified from the data. The following node labels were used:

- Pipe
- Pump
- Filter
- Refiner
- Screen
- Cleaner
- Cylinder

- Air compressor
- Heat exchanger/condenser
- Fan
- Large tank
- Small tank
- Tank of unknown size

Edges represent a connection between two components. The direction of the edge specifies the direction of the flow and the edge label the type of the flow. Flow types are represented by high-level flow groups. The used flow groups are:

- Additives
- Broke
- Chemicals
- Condensate
- Effluent
- Gas
- Pulp
- Sludges and pigments
- Steam
- Vapor
- Water
- White water

As with the node types, the flow group can also be unknown. In addition, one edge is allowed to have multiple flow groups.

A simplified example of a process graph is shown in Figure 4. Colors are used to denote the flow group of each edge, with red corresponding to pulp and brown to broke. Information about the process graphs of each site has been collected to Table 2.

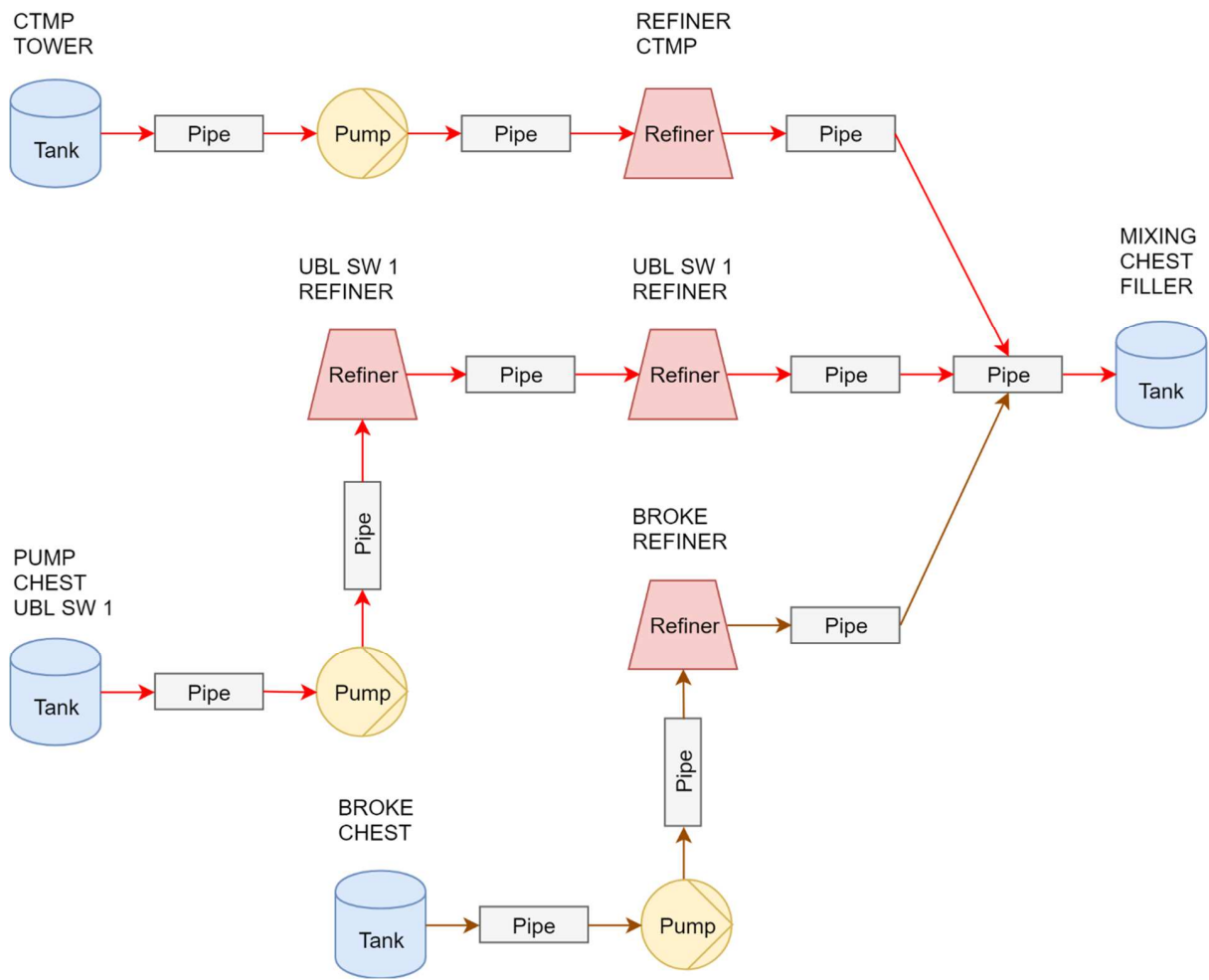

*Figure 4 Example process graph structure*

*Table 2 Summary of process graphs*

|  | Site A | Site B | Site C | Site D | Early site A |
|---|---|---|---|---|---|
| **Nodes** | 5544 | 4587 | 1794 | 2312 | 879 |
| **Edges** | 5977 | 5911 | 2284 | 3177 | 780 |
| **Different node types** | 8 | 10 | 8 | 10 | 7 |
| **Different edge types** | 12 | 10 | 11 | 10 | 6 |

### 4.1.2 Required source information for graph building

As has been described in section 3, the source data can be in varying formats, for example, a 3D plant design or a spreadsheet. This subsection highlights the data that needs to be extracted. Two tables are needed to define one graph: a node table for specifying the components and an edge table for specifying the connections between the components. The node table should include a unique id for every node and also the type of the node; an example is presented in Table 3. The edge table should include an edge id, the node id of the source node, the node id of the target node and the flow group of the edge; an example is presented in Table 4. These two tables should be available for both the source and the target model graphs in order to perform the matching. They are the source information for the next step of the methodology described in section 4.2.

*Table 3 Node table example*

| Node id | Node type |
| --- | --- |
| Node1 | Pump |
| Node2 | Pipe |
| Node3 | Unknown |
| Node4 | Refiner |
| Node5 | Pipe |
| … | … |

*Table 4 Edge table example*

| Edge id | Source node | Target node | Flow group |
| --- | --- | --- | --- |
| Edge1 | Node1 | Node2 | Water |
| Edge2 | Node2 | Node3 | Water |
| Edge3 | Node4 | Node5 | Pulp |
| … | … | … | … |

## 4.2 Graph simplification

Graph simplification is desirable for the following reasons:

1. The execution times and memory requirements of the graph matching algorithms depend on the number of nodes in the graph
2. In the original graphs, most of the nodes were pipes. In the graphs created from the 3D data, over 80% of nodes were pipes. Without simplification, other types of nodes would only be connected to pipe nodes, so it is difficult to use the immediate neighborhoods to identify the nodes themselves.
3. A single pipe node representing a pipe line in the line list data could be broken into several pipe nodes in the 3D data that represented the branches of the same pipe line. Simplifying the piping makes the graphs created from different sources more similar to each other. This reduces the effect that the data source has on the graph and thus improves the generality of the solution.

The following proposed simplification algorithms reduce the number of pipe nodes. Any valves or other instrumentation on the removed pipe sections are also removed. Their impact will be evaluated empirically.

### 4.2.1 Removal of pipes with degree 1

The graphs had a large number of pipe nodes with degree 1, i.e. nodes that were connected to only one edge. The pseudocode for removing them is presented below. First, all the pipe nodes with degree 1 are collected and removed. This is repeated until the graph no longer has any pipe nodes with degree 1. An example is presented in Figure 5 Iterative removal of pipe nodes with degree 1.

**Algorithm 1** Removal of pipe nodes with degree 1

1: **procedure** removePipesWithDeg1(Graph G)

2:     pipeNodeList ← G.getNodesWithType("Pipe");

3:     **repeat**

4:         oldSize ← G.size();

5:         **for each** (node n in pipeNodeList) **do**

6:             **if** (n.getDegree() == 1) **then**

7:                 pipeNodeList.remove(n);

8:                 remove(n); // Removes node and its edges from graph

9:         newSize ← G.size();

10:    **until** (oldSize == newSize)

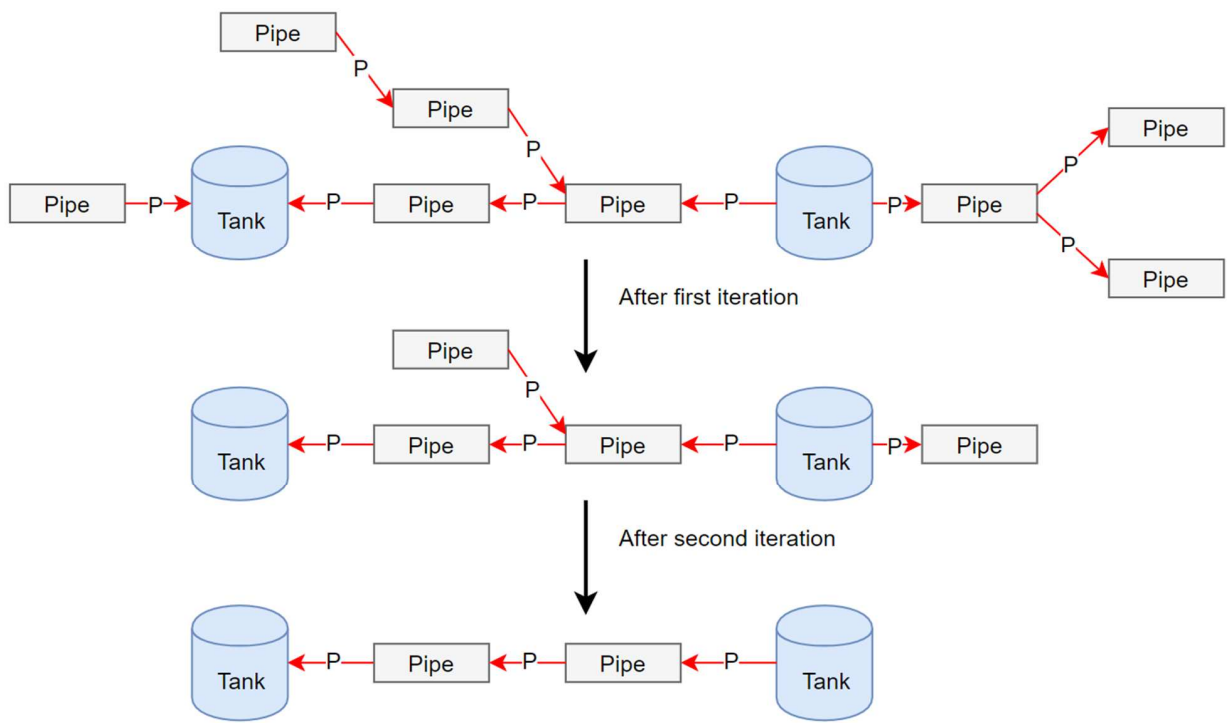

*Figure 5 Iterative removal of pipe nodes with degree 1*

### 4.2.2 Removal of straight pipelines

The following pseudocode removes all pipe nodes that have exactly one incoming and one outgoing edge. An example is provided in Figure 6 under the label "Simplification of straight pipelines:".

**Algorithm 2** Simplification of straight pipe lines

1: **procedure** simplifyStraightPipelines(Graph G)

2:     pipeNodeList ← G.getNodesWithType("Pipe");

3:     **for each** (node n in pipeNodeList) **do**

4:             edgesOut ← n.outgoingEdges();

5:             edgesIn ← n.incomingEdges();

6:             **if** (edgesOut.length == 1 AND edgesIn.length == 1) **then**

7:                     eIn ← edgesIn[0];

8:                     eOut ← edgesOut[0];

9:                     nIn ← eIn.startNode();

10:                    nOut ← eOut.endNode();

11:                    flowGroups ← eIn.flowGroups() + eOut.flowGroups();

12:                    createNewEdge(nIn, nOut, flowGroups);

13:                    remove(n);

### 4.2.3 Removal of all pipes

Final simplification algorithm removes all pipe nodes from the graph and replaces them with edges, leaving only non-pipe nodes in the graph. In the simplified graph, an edge connecting two nodes means that a directed path through pipe nodes existed between those nodes in the original graph. The new edge contains all of the flow groups found on the original path. The pseudocode is presented below and an example is provided in Figure 6 under the label "Removal of all pipes:".

**Algorithm 3** Removal of all pipes

1: **procedure** removeAllPipes(Graph G)

2:     pipeNodeList ← G.nodesWithType("Pipe");

3:     **for each** (node n in pipeNodeList) **do**

4:             edgesOut ← n.outgoingEdges();

5:             edgesIn ← n.incomingEdges();

6:             **if** (edgesOut.length > 0 AND edgesIn.length > 0) **then**

7:                     **for each** (edge eIn in edgesIn) **do**

8:                             **for each** (edge eOut in edgesOut) **do**

9:                                     nIn ← eIn.startNode();

10:                                    nOut ← eOut.endNode();

11:        flowGroups ← eIn.flowGroups() + eOut.flowGroups();

12:        createNewEdge(nIn, nOut, flowGroups);

13:     remove(n);

**Original:**

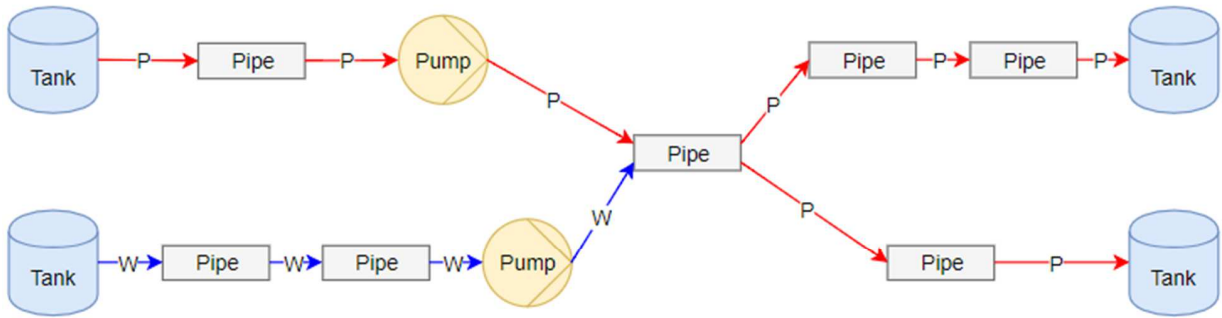

**Simplification of straight pipelines:**

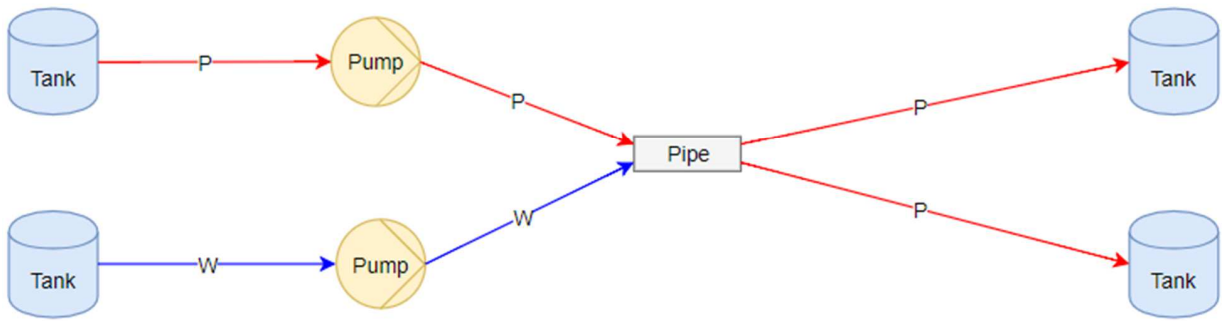

**Removal of all pipes:**

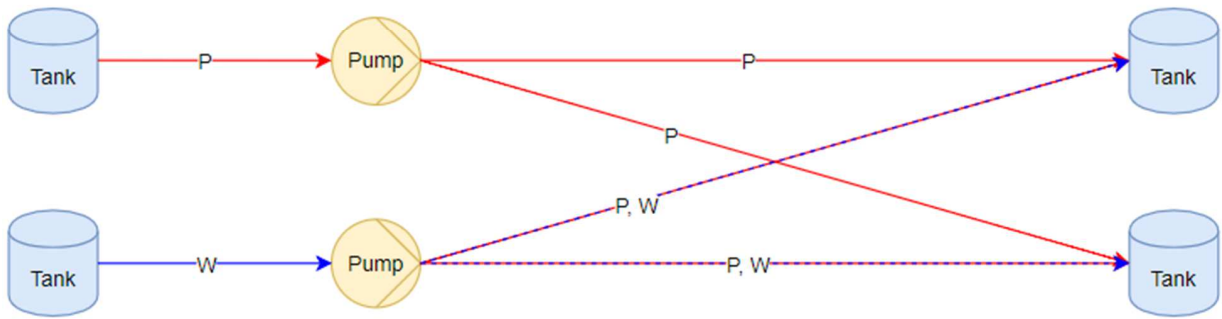

*Figure 6 Comparison for algorithm 2 (simplification of straight pipelines) and algorithm 3 (removal of all pipes)*

Algorithms 2 and 3 can create multi-edges. These are edges that have the same start and end nodes, but can otherwise be different, for example by containing different flow groups. Graph matching algorithms cannot usually handle multi-edges, which means that they have to be combined into regular edges. This can be done using the following procedure illustrated in Figure 7:

1. Find edges that share the same start node and end node.
2. Collect the flow groups of these edges.
3. Combine these edges into a new edge that has the start node, end node and the all of the flow groups of the combined edges.

**Original:**

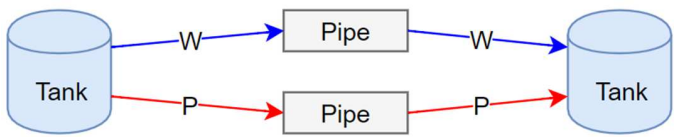

**After simplification:**

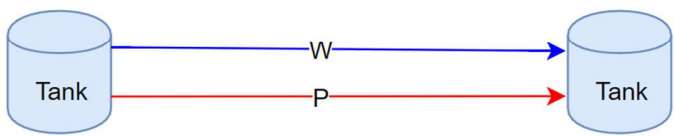

**After combining multi-edges:**

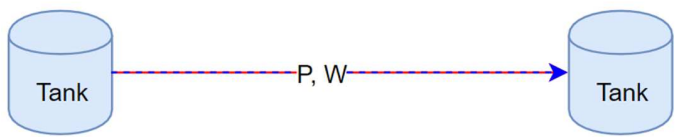

*Figure 7 Combination of multiple parallel edges after simplification*

## 4.3 Node similarity measures

As was discussed in section 2, a new similarity measure needs to be developed for the process plant domain. Similarity is described with a value from 0 to 1, where a larger value implies a greater similarity. Four alternatives are proposed below. These will be evaluated empirically.

Comparison of node types can be used to define a **node type similarity measure** according to Table 5. There was no analytical way of determining a specific similarity value for cases involving an unknown type, and the practical value of our proposal will be evaluated empirically.

*Table 5 Node type similarity scores*

| Compared node types | Example | Similarity |
|---|---|---|
| Two known same node types | Refiner - Refiner | 1 |
| Two known different node types | Refiner - Pump | 0 |
| Unknown and known type | Unknown - Refiner | 0.5 |
| Two unknown node types | Unknown – Unknown | 0.8 |
| Two tanks of same size class | Large tank - Large tank | 1 |
| Two tanks of different size class | Large tank - Small tank | 0 |
| Unknown size tank with any tank | Unknown size tank - Large tank | 0.8 |

The **neighborhood similarity measure** is adapted from [51] as follows. Each neighbor is described with a pair (nodeType, edgeType), and two neighbors are considered to be similar if they have both a matching nodeType and edgeType as specified in Table 6 and Table 7. Edge direction is considered by only allowing in-neighbors to be matched with in-neighbors and out-neighbors with out-neighbors. Once the number of similar neighbors has been found, it is divided by the degree of the source graph node to obtain the final similarity value.

Figure 8 shows two example graphs to be matched. Table 8 has a comparison of in-neighbors of the central node, i.e. nodes from which there is an edge into the direction of the central node. The in-neighbors of the graph on the left are in rows and those of the graph on the right are in the columns. Similarly Table 9 has a comparison of the out-neighbors. The assignment problem matrices in Table 6 Table 7 are given to the Hungarian algorithm [71]. In the in-neighbor case, (Pump, W) would be matched with (Pump, W), (Refiner, Unknown) with (Refiner, P), (Filter, P) with (Unknown, P) and the remaining (Pump, W) would remain unmatched. In the out-neighbor case, (Tank, [P, W]) would be matched with (Tank, W) and (Tank, P) with (Tank, P). Therefore, the total number of matching neighbors between the two nodes is 5. As the degree of the first node is 6, the final similarity would be 5/6.

*Table 6 Equal neighbors, node type comparison*

| Compared node types | Example | Match |
|---|---|---|
| Two known same node types | Refiner - Refiner | x |
| Two known different node types | Refiner – Pump | |
| Unknown and known type | Unknown - Refiner | x |
| Two unknown node types | Unknown - Unknown | x |

*Table 7 Equal neighbors, edge type comparison*

| Compared edge types | Example | Match |
|---|---|---|
| Two known same edge types | Water - Water | x |
| Two known different edge types | Water – Pulp | |
| At least one shared type | Pulp, water - Pulp, white water | x |
| No shared edge types | Pulp, broke - Steam, condensate | |
| Known and unknown edge type | Water – Unknown | x |
| Two unknown edge types | Unknown - Unknown | x |

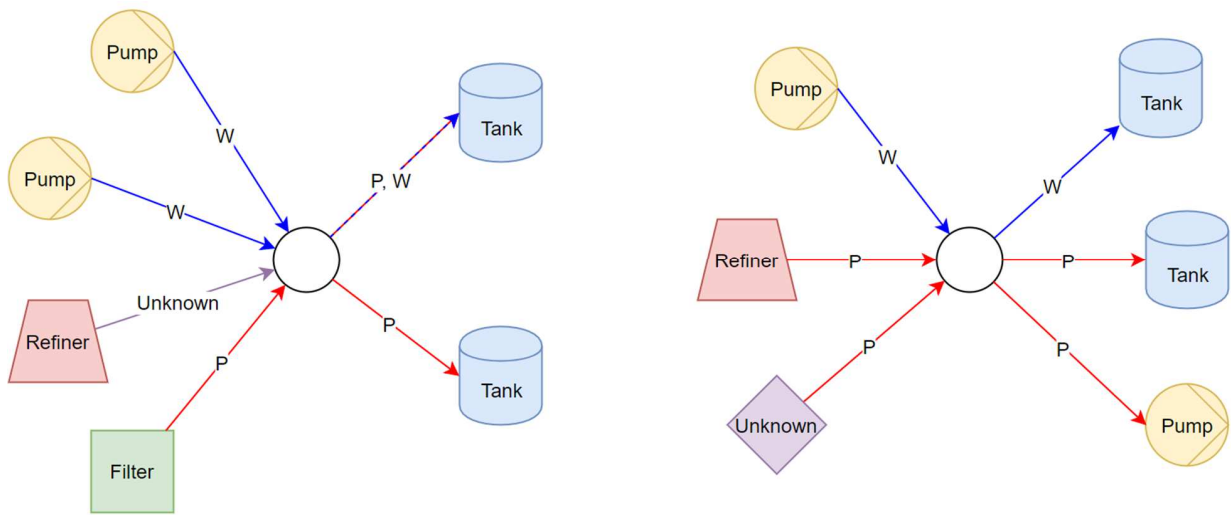

*Figure 8 Neighborhood similarity example*

*Table 8 Comparison of in-neighbors of Figure 8. Value 1 indicates that the neighbors are considered a match.*

|  | **(Pump, W)** | **(Refiner, P)** | **(Unknown, P)** |
|---|---|---|---|
| **(Pump, W)** | 1 | 0 | 0 |
| **(Pump, W)** | 1 | 0 | 0 |
| **(Refiner, Unknown)** | 0 | 1 | 1 |
| **(Filter, P)** | 0 | 0 | 1 |

*Table 9 Comparison of out-neighbors of Figure 8. Value 1 indicates that the neighbors are considered a match*

|  | **(Tank, W)** | **(Tank, P)** | **(Pump, P)** |
|---|---|---|---|
| **(Tank, [P, W])** | 1 | 1 | 0 |
| **(Tank, P)** | 0 | 1 | 0 |

The **anchor similarity measure** uses anchor pairs, i.e. a pair of nodes from the two graphs to be compared which the user has manually marked as being similar. These are useful in process graphs with repeating subsections, since differentiating between these sections is difficult based on the graph structure alone. If an anchor is chosen from each subsection, that information can help the graph matching algorithms to match the rest of the nodes to the correct sub-sections as well. The position of a node with respect to the anchor nodes is measured by calculating the shortest paths between the node and all of the anchor nodes. As the graph is directed, two distances can be calculated for each anchor: the shortest path downstream and the shortest path upstream. If a path does not exist, the path length is set to infinity. The shortest paths can be calculated with Dijkstra's algorithm [72]. The path lengths are collected into an anchor vector. This anchor vector is calculated for every node in both graphs and the similarity between two nodes can be calculated by comparing their anchor vectors. The anchor vectors have the following elements in this order:

- Shortest path from the first anchor downstream
- Shortest path from the first anchor upstream
- Shortest path from the second anchor downstream
- Shortest path from the second anchor upstream

An example is shown in Figure 9, in which the anchor vectors for nodes $v_1$ and $v_2$ are shown.

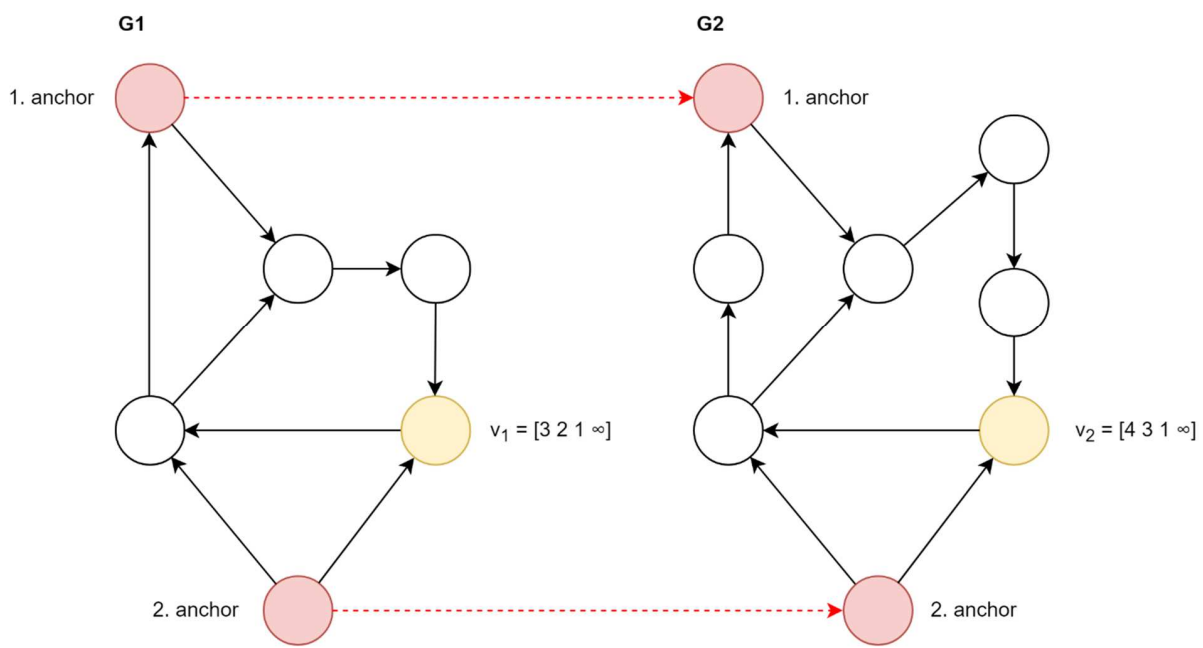

*Figure 9 Anchor similarity example. Anchor vectors, with respect to the red anchor nodes, have been calculated for the yellow nodes.*

If a node is located far away from an anchor pair, the possibility increases that broken connections or design differences, such as 2 components connected in series instead of 3, affect the length of the shortest path between the node and the anchor. Thus, two nodes with path lengths 15 and 20 to a shared anchor are more likely to be a match than two nodes with path lengths 1 and 6. Instead of comparing the shortest path lengths directly, Equation 1 is proposed to compute the *influence* that each anchor has on a node.

*Equation 1 Influence of an anchor on a node. x is path length from anchor to node*

$$r(x) = e^{-\frac{x}{\sigma_1}}$$

Equation 1 is applied to each element of the anchor vectors $v_1$ and $v_2$, which are then given as input to Equation 2:

*Equation 2 Gaussian similarity function to calculate the similarity between the two vectors*

$$w(v_1, v_2) = e^{-\frac{\|v_1 - v_2\|^2}{\sigma_2}}$$

The empirical evaluation in section 6 uses $\sigma_1 = 5$ and $\sigma_2 = 1$. Equation 2 returns a value between 0 and 1 where a value closer to 1 implies a greater similarity between the two vectors.

Most graph matching algorithm allow only one similarity score for each node pair. Equation 3 combines the three measures proposed above into a single measure. The weight *a* is in the range [0,1] and can be determined empirically.

*Equation 3 Formula for combining our three node similarity measures*

*(a · neighbourScore + (1 – a) · anchorScore) · nodeTypeScore*

## 4.4 Exploiting the results

This subsection will discuss the output of the methodology and the exploitation of this output with respect to the use case in section 1.

The result of the matching shows a corresponding node in the target graph for every node in the source graph. An example is provided in Table 10:

*Table 10 Matching results table*

| Source graph node | Target graph node |
| --- | --- |
| sNode1 | tNode3 |
| sNode2 | tNode2 |
| sNode3 | tNode6 |
| sNode4 | tNode1 |
| sNode5 | tNode4 |
| … | … |

For example, when choosing parameters for the component corresponding to node sNode3 in the source graph, the component corresponding to node tNode6 in the target graph could be used as a reference. This information about corresponding components can provide support in the initial parameter estimation use case. Two components that perform a similar function frequently have similar parameters. Therefore, information about corresponding components in previous projects can provide guidance to the designer, as he or she is selecting initial component parameters for the new project. This information could also serve as a basis for an automated solution in further work. Such a solution would find corresponding components automatically from previous projects and import their parameters values to the components of the new project.

# 5 Illustrative example

In this section, the methodology presented in section 4 is illustrated for the multi-board plant subsection in Figure 1. Figure 4 showed a simple example process graph, which formed the starting point of the methodology. The corresponding graph for the plant in Figure 1 is presented in Figure 10.

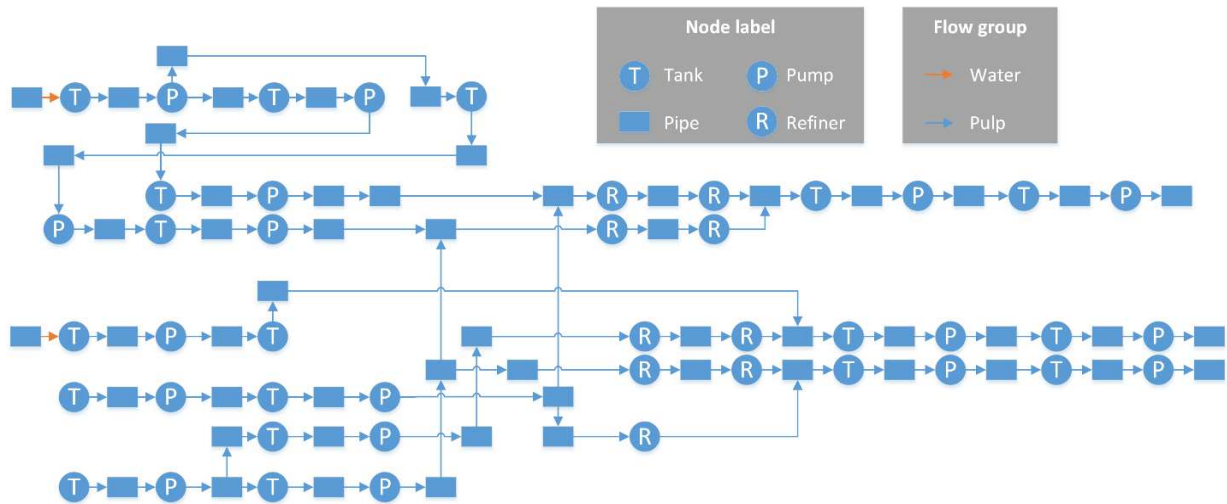

*Figure 10 Process graph for the plant in Figure 1*

In Figure 5, all pipelines of degree 1 were removed after two iterations of Algorithm 1. Figure 11 shows the result of applying Algorithm 1 to Figure 10; in this case the algorithm terminated after one iteration. Only a few pipe nodes at the far left and right of Figure 10 were removed.

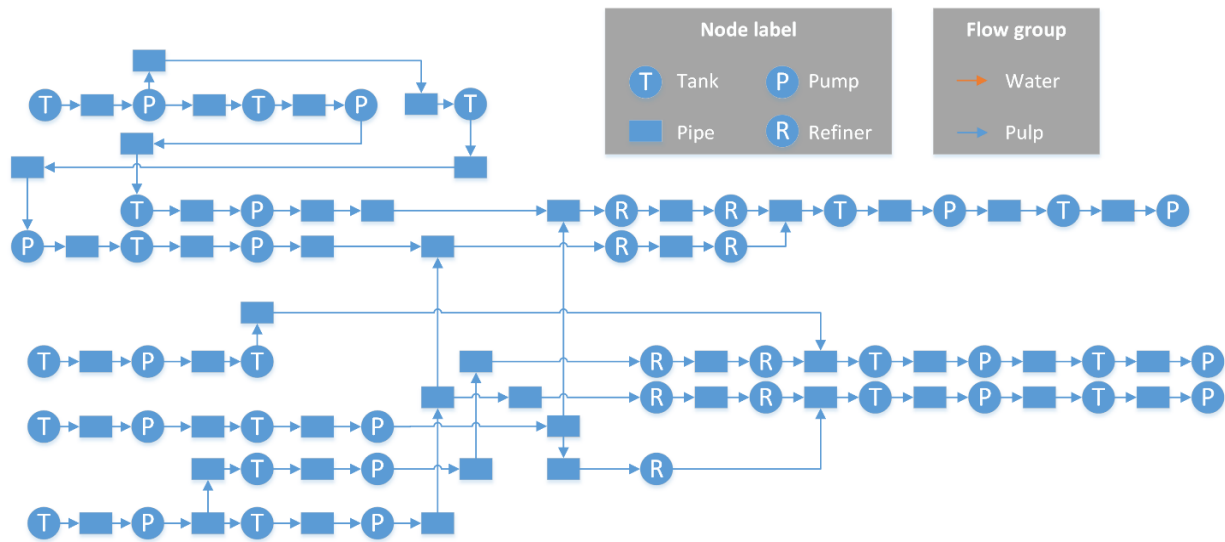

*Figure 11 Process graph from Figure 10 after removal of pipe nodes with degree one*

The result of applying Algorithm 2 to Figure 11 is presented in Figure 12. This stage corresponds to the middle graph in Figure 6.

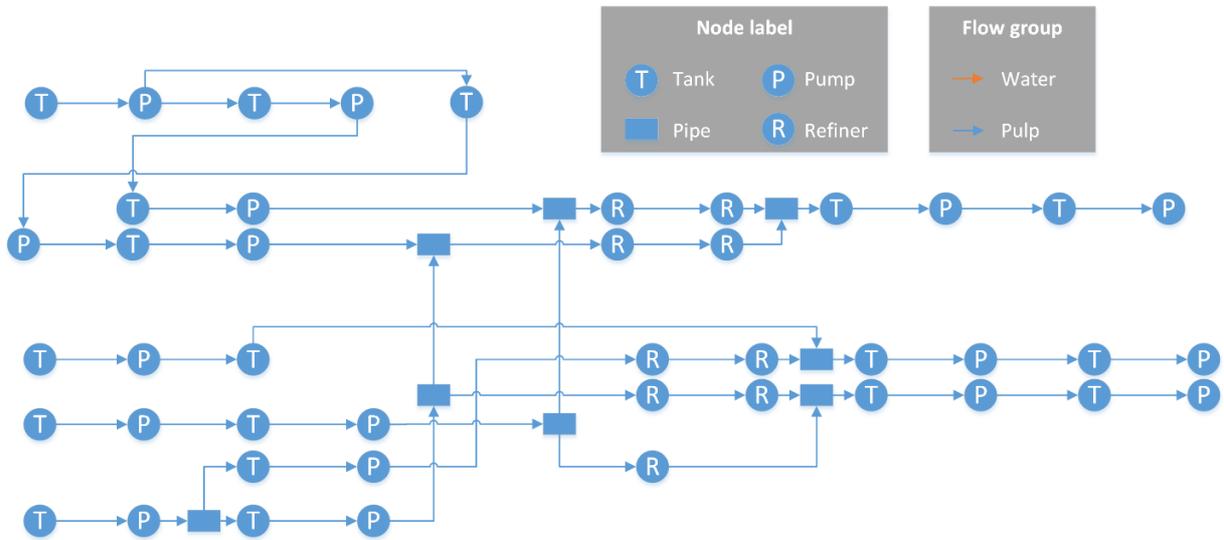

*Figure 12 Graph from Figure 11 after removing all straight pipes*

Figure 13 shows the result of applying Algorithm 3 to the graph in Figure 12. This stage corresponds to the bottom graph in Figure 6. Algorithm 3 could also have been applied directly to the original graph in Figure 10, in which case the outcome would have been the same as depicted in Figure 13.

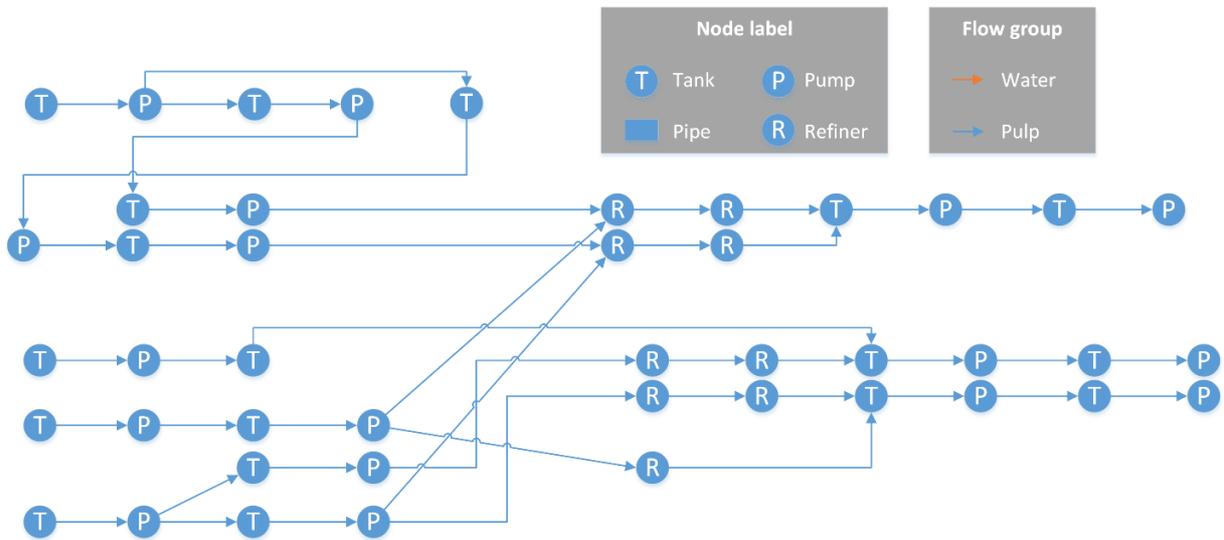

*Figure 13 Graph from Figure 12 after removal of all pipes*

With regard to the multi-edge combination illustrated in Figure 7, there is no impact to the graph in Figure 13.

# 6 Experiment description

## 6.1 Definition of experiments

Graph matching was performed between 3 different site pairs:

- Site A - Site B
- Early site A - Site B
- Site C - Site D

Compared simplification algorithms (see section 4.1) were:

- Simplification 1: Removal of 1-degree pipes + removal of straight pipelines
- Simplification 2: Removal of all pipes

Table 11 shows the impact of applying these algorithms.

*Table 11 Number of nodes left in the graphs after applying the simplification algorithm*

|                  | Site A | Site B | Site C | Site D | Early site A |
|------------------|--------|--------|--------|--------|--------------|
| **Original**         | 5544   | 4587   | 1794   | 2312   | 879          |
| **Simplification 1** | 1138   | 888    | 938    | 1127   | 101          |
| **Simplification 2** | 606    | 462    | 680    | 855    | 32           |

Compared node similarity measures (see section 4.2) are:

- Similarity function 1: Only node type similarity is used
- Similarity function 2: Neighbor + node type similarity (set a = 1 in Equation 3)
- Similarity function 3: Anchor + node type similarity (set a = 0 in Equation 3)
- Similarity function 4: Weighted combination of neighbor and anchor similarity with node type similarity (set a = 0.5 in Equation 3)

Compared graph matching algorithms are:

- DSPFP 1-to-1
- DSPFP M-to-1
- Similarity flooding
- WAVE
- GEDEVO
- SANA

An experiment was performed for every combination of site pairs, simplification algorithms, node similarity functions and graph matching algorithms as follows:

1. Choose two process plant sites to be matched. Create graphs for this pair of sites, noted by G1 and G2.
2. Select a simplification algorithm and apply it to both G1 and G2. The result is two simplified graphs, simpleG1 and simpleG2.

3. Select a node similarity function and use it to calculate the node-to-node similarity matrix between the graphs simpleG1 and simpleG2.
4. Select a graph matching algorithm, and give the graphs simpleG1, simpleG2 and the node-to-node similarity matrix as an input to the algorithm.
5. Run the graph matching algorithm and calculate the percentage of correct matches with respect to the handpicked matches.

This resulted in 114 experiments. Each experiment was run 20 times with maximum runtime 5min. 3 anchors were used for the site A- site B and early site A - site B matching. 2 anchors were used for site C -site D matching. Mixing tanks in the middle of each repeating area were chosen as the anchor nodes.

## 6.2 Implementation and parameterization of algorithms

### 6.2.1 DSPFP

DSPFP had two main parameters to be tuned: the step size $\alpha$ and the weight of the node similarity $\lambda$. As mentioned in the original article, the matching quality is not very sensitive to the parameter $\lambda$ [44]. As changing $\lambda$ or $\alpha$ did not have a considerable effect on the results, the same values $\alpha = 1$ and $\lambda = 0.5$ were used during the final testing as in the original article.

As the algorithm was self-implemented in Java, additional parameters such as the convergence criteria had to be set. The algorithm consists of an inner and outer loop, which are both repeated until the matrix they operate on converges. For both loops, the same convergence limit 0.00001 was used. The error was calculated by subtracting the new matrix from the old one, taking the absolute value of each element of the resulting matrix and then calculating the total sum of all the elements. In addition, the maximum number of iterations was set to 100 for both loops. The loops were terminated if either the error fell below the convergence limit or the number of iterations exceeded the maximum value.

During the discretizaion phase, the same one-to-one greedy algorithm of [42] was used as in the original DSPFP article. However, in addition to the one-to-one algorithm, another many-to-one greedy algorithm was specified [42]. To test if the many-to-one algorithm could provide better results, it was implemented alongside the one-to-one algorithm used in the original article. In this article, the basic one-to-one version of DSPFP is referred to as DSPFP 1-to-1 and the many-to-one version as DSPFP M-to-1.

## 6.2.2 Similarity flooding

The similarity flooding article provided many customization options for the algorithm. They can be divided into three categories: the propagation coefficients, the fix-point formulas and the filters. The propagation coefficient is defined with a function that calculates how the similarity scores of the node pairs propagate to their neighbours. In the extended version [73] of the similarity flooding article [34], seven options for calculating the propagation coefficient were presented. As the inverse product formula provided the best results in the article, it was also the one implemented in this thesis. For the fix-point formulas that increment the similarity scores during each iteration, four different options were presented: A, B, C and basic. In the article, B and C provided the best results but formula C converged considerably faster than the others. Formula C was also chosen in this thesis. Many options for the filters that determine the type of the mapping were also presented in the article. The many-to-many filter "selectThreshold" performed best in the article and was also implemented in this thesis. The same threshold value $t_{rel}$ = 1.0 was selected as in the evaluation section of the article. The article also recommended using application dependant constraints to filter out matches based on the node types. Accordingly, the final results were also filtered so that matches between different node types were removed.

### 6.2.3 WAVE

For the WAVE algorithm, only the node similarity scores had to be given as an input for the calculation of WEC and WNC. The algorithm did not have any additional configurable parameters.

### 6.2.4 GEDEVO

GEDEVO allows the user to set the costs of the GED operations. These operations also included options for directed graphs, where a cost could be assigned for flipping the direction of an edge. While the other costs were kept at their default values, the penalties for flipping an edge direction or changing an directed edge into an undirected one were set to 0.95 on a scale of 0 to 1.

To calculate the pair score weights, a combination of GED and graphlets was used as a default. The relative weight of GED was 1 and the relative weight of graphlets was 0.5. In order to include the node similarity scores, a new weight distribution needed to be found. This was done by setting the weight of the node similarity score to 0.5 and testing different weight combinations for the GED and the graphlets. All the tested combinations are shown in Table 12, from which GED = 0.4, graphlet = 0.1 and node similarity = 0.5 produced the best results and were chosen for final testing.

*Table 12 GEDEVO pair score parametrization*

| GED | Graphlet | Node similarity |
|-----|----------|-----------------|
| 0.5 | 0 | 0.5 |
| 0.4 | 0.1 | 0.5 |
| 0.3 | 0.2 | 0.5 |
| 0.2 | 0.3 | 0.5 |
| 0.1 | 0.4 | 0.5 |
| 0 | 0.5 | 0.5 |

### 6.2.5 SANA

The implementation of the SANA algorithm allows the user to choose from many different objective functions, such as alignment quality measures EC, $S^3$ and WEC described in Chapter 2. As the default objective function, SANA optimizes the $S^3$ metric. Of the provided objective functions, $S^3$, EC, graphlets and WEC were tested with the node similarity scores. WEC was set to use the node similarity scores to calculate the edge weights. To find out the most suitable objective function, the weight of the node similarity scores was set to 0.5, and the other objectives were tried one-by-one with a weight 0.5. Of the different objective functions, EC and WEC returned the best results. There was not a considerable difference in the results when the weights were 0.5, but if only EC = 1 or WEC = 1 were used with the node similarity score weight set to 0, the number of matches found with EC was close to zero, while WEC continued to return good results. This is likely because WEC also includes the node similarities in the edge weights. In the end, WEC was chosen as the more robust option.

## 6.3 Evaluation of results

To evaluate the graph matching results, a number of correct matches were handpicked from the graphs. 111 matches were found for the site A-B comparison and 113 for the site C-D comparison. For the comparison between the early site A and site B, 25 matches were used. As finding matches manually was a time consuming task and the graph sizes were large, these matches did not represent all the possible matches that existed between the two graphs. The only exception was the comparison between early site A and site B, where the graph size of early site A was small enough so that all matches could be found. The evaluation of empirical results was done by calculating the percentage of correctly identified matches. The performance was described by recall, which was the percentage of handpicked matches that the algorithm matched correctly. The one-to-one algorithms could not even theoretically achieve 100% recall as there were cases where 2 or more nodes of the source graph should have been matched to the one same node in the target graph.

# 7. Results

The results for sites A and B are presented in Figure 14. As discussed in section 3.1, both of these sites are multi-board plants, and a human expert would recognize that they have similar process structure with several matching process components, so the purpose of the experiments in Figure 14 is to assess the algorithms capability to find these matches. The recall is the percentage of these matches that were found by the algorithm, as explain in more detail in section 4.3. Figure 15 repeats the experiment for early site A and site B. This corresponds to a scenario in which there is only an early phase design (site A) being matched to a detailed previous design (site B). Figure 16 presents the results for the matching of sites C and D. These experiments are otherwise similar to those presented in Figure 14, the difference being that sites C and D are recycled fiber plants.

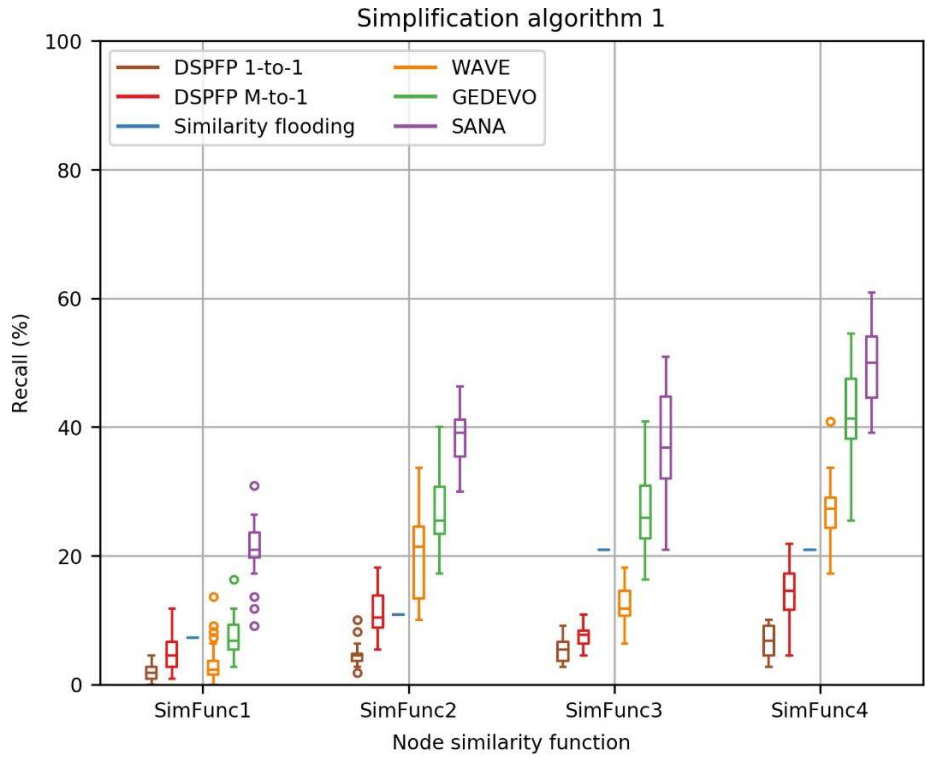

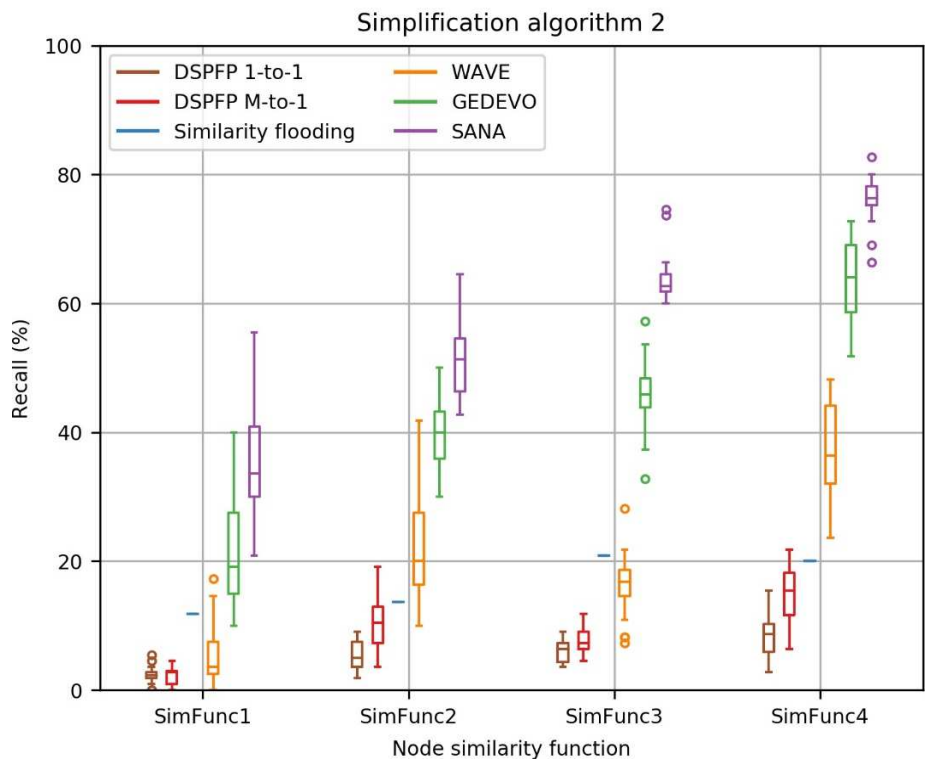

*Figure 14 Graph matching between sites A and B, showing recall of matches to exact counterparts*

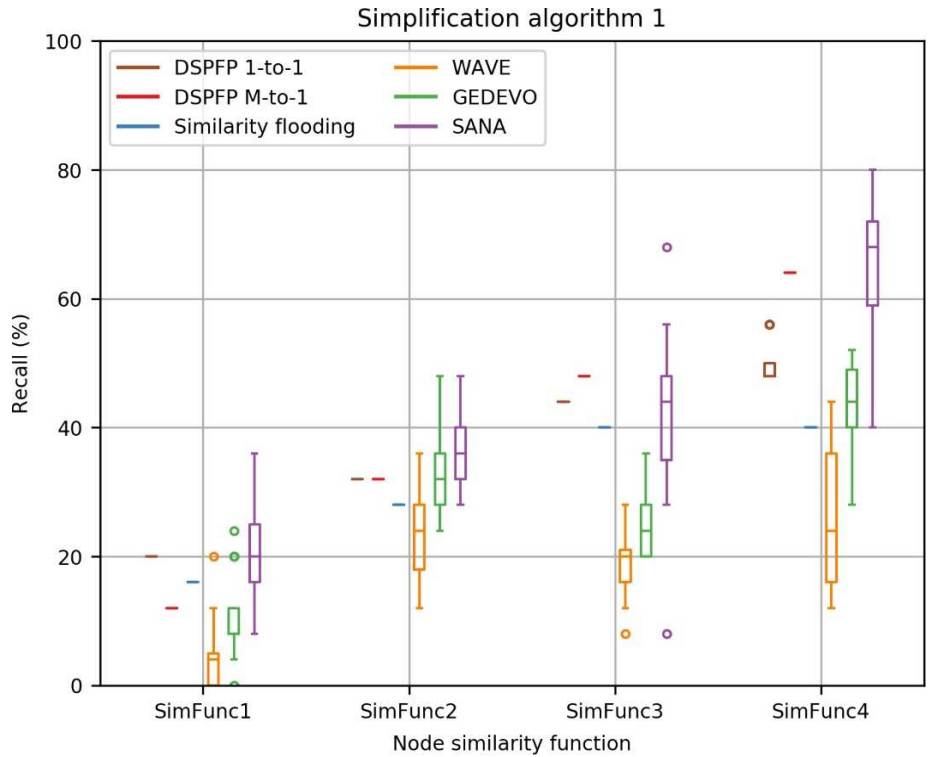

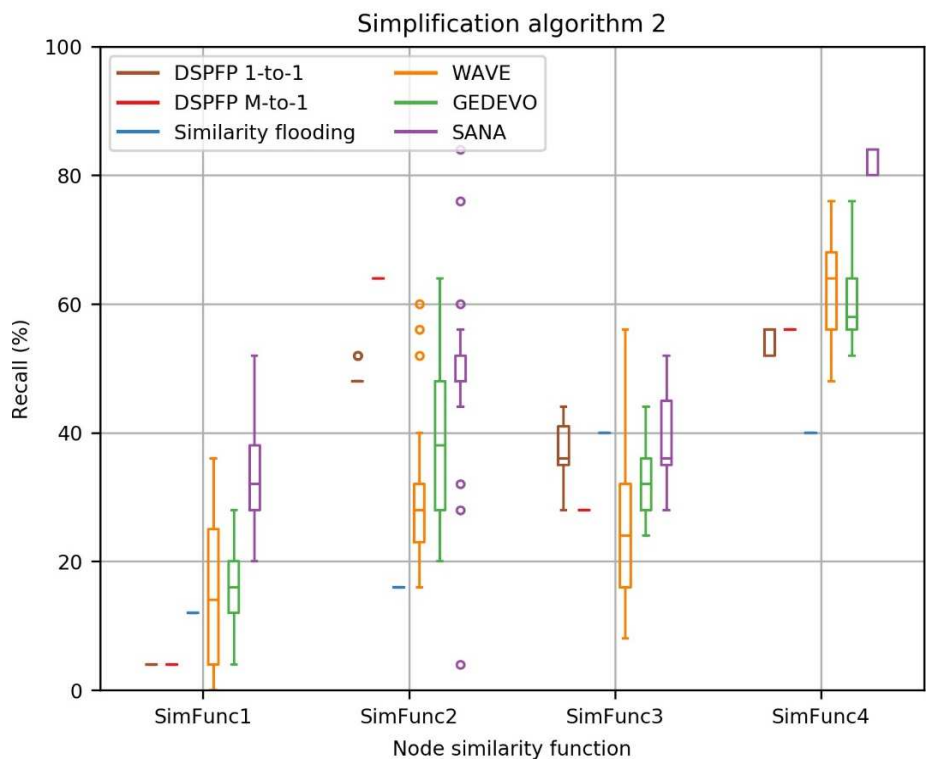

*Figure 15 Graph matching between early site A and site B, showing recall of matches to exact counterparts*

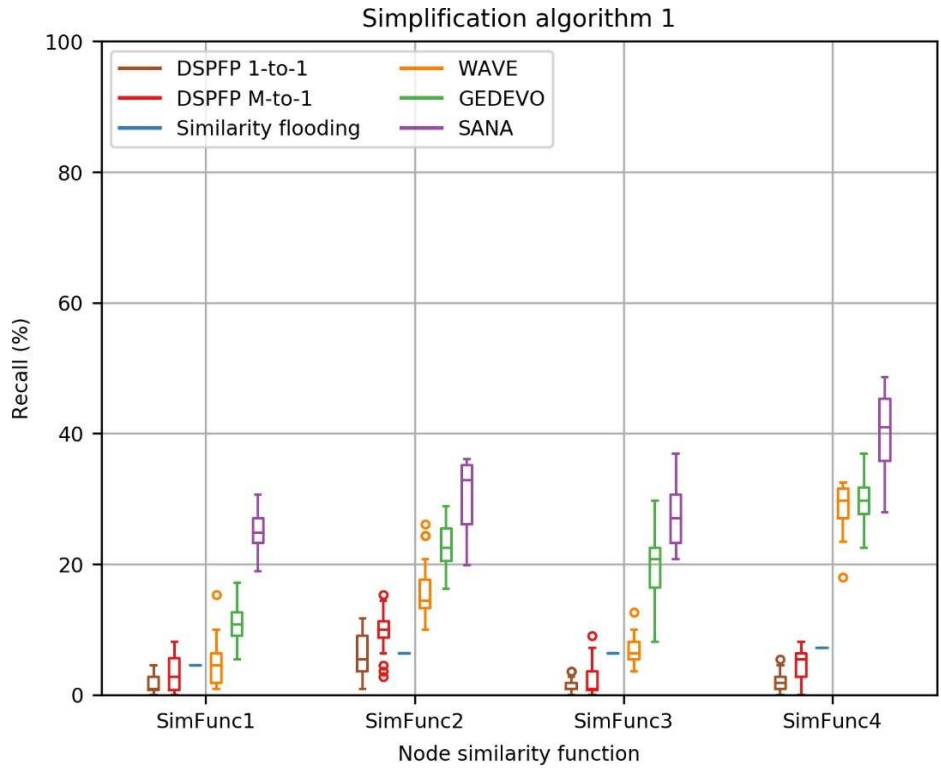

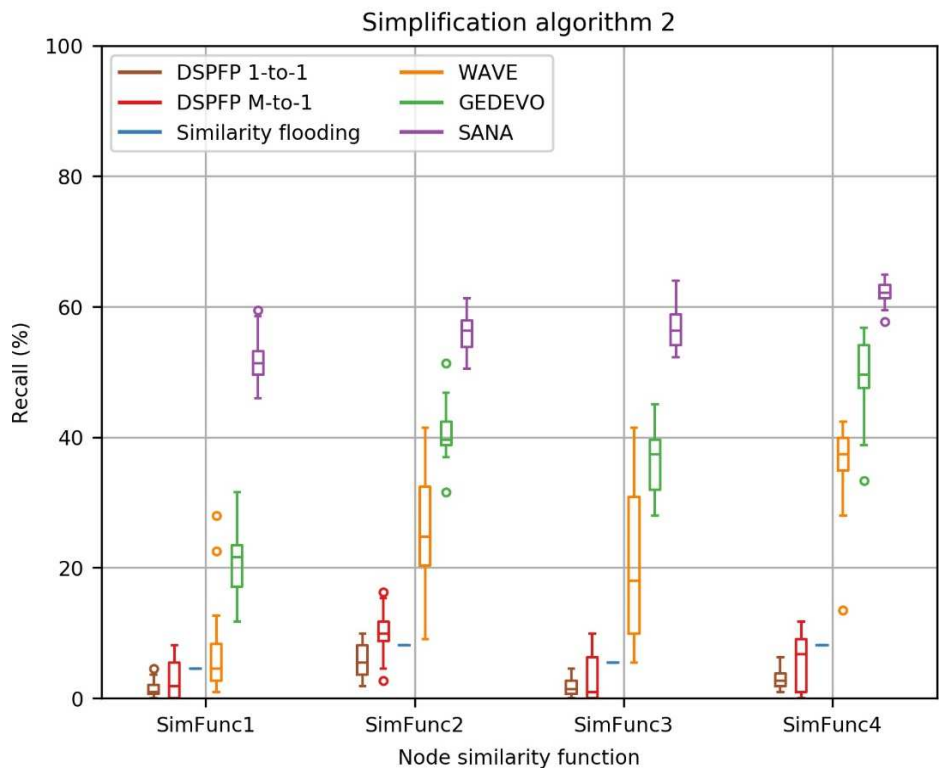

*Figure 16 Graph matching between sites C and D, showing recall of matches to exact counterparts*

The best combination was SANA with simplification function 2 and node similarity function 4, so further experiments were made on it as follows.

Many graph matching algorithms match every node of the smaller graph with a node from the larger graph. This is not a desirable result if the graphs contain outliers, in other words nodes that do not have a match in the other graph. For example, the graphs for sites A and B contained many nodes that did not have a match in the other graph. This is because both sites contained areas that were not part of the scope of the other site. If all nodes are matched, it becomes difficult to find the actual matches from all of the returned results. Some kind of filtering is therefore needed get rid of these false matches. For example, the node similarity scores can be used here. Higher similarity represents a higher degree of confidence that the match is correct. Node similarity between the two matched nodes can be calculated and the match can be discarded if the value is below a chosen limit.

The node similarity scores of the matched pairs were used to filter out incorrect matches, in which node similarity was below a limit value. As the correct correspondences were known for only a set of manually matched nodes, the filtering was evaluated by investigating how many matches can be removed before there is a significant effect on the recall. The results in Figure 17 suggest that a good filter limit for our use case should be in 0.5 - 0.6 range.

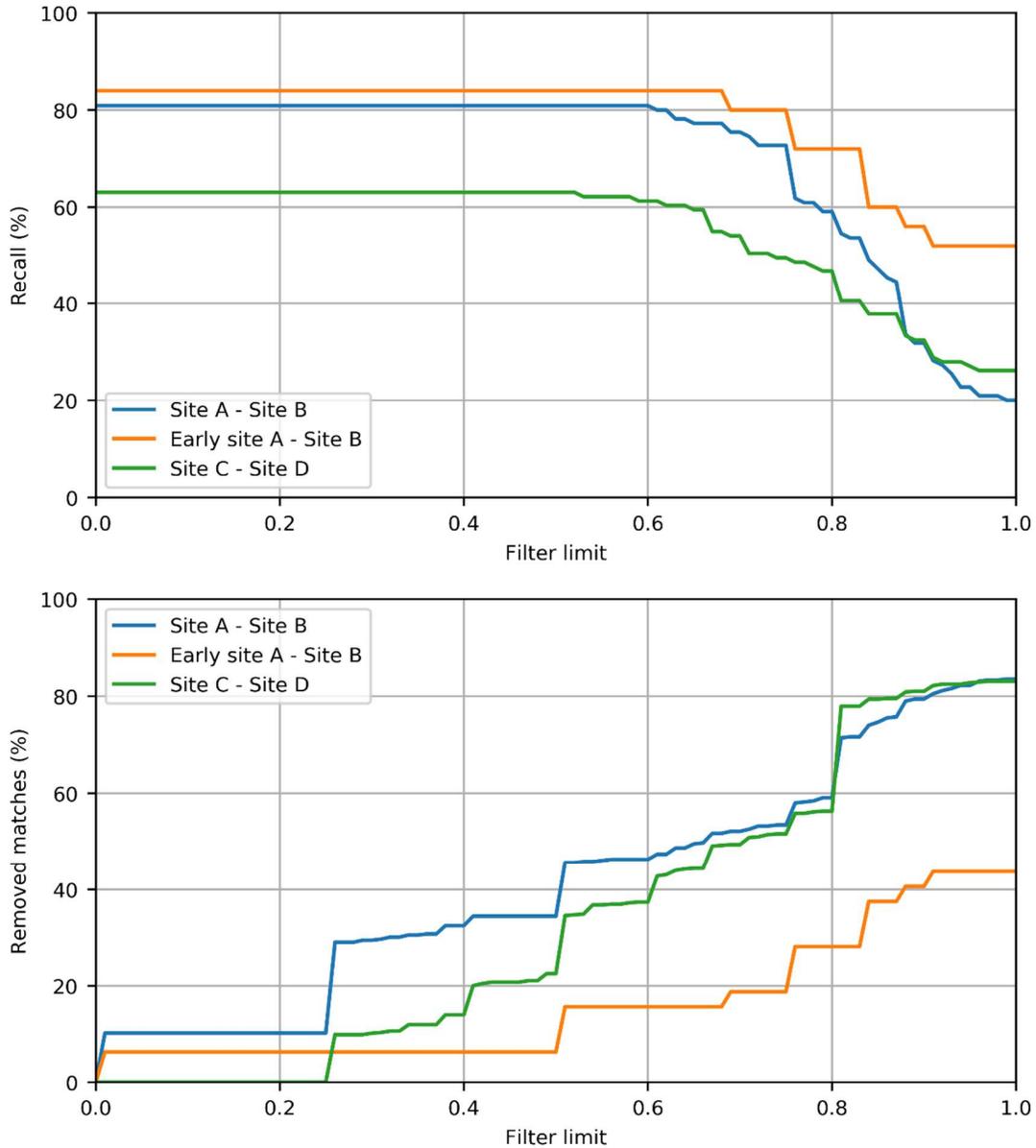

*Figure 17 Filtering matches using node similarity function 4 (see section 5.1) scores as the filter limit*

The termination criterion for the best performing algorithm SANA was its runtime. In Figure 14 - Figure 16, the runtime was set to 5 minutes. The best performing combination was tested with different runtimes from 0.1 to 30 minutes. As can be seen from Figure 18, the algorithm converges in 2min for site A - site B and site C - site D and in 0.3min for early site A – site B.

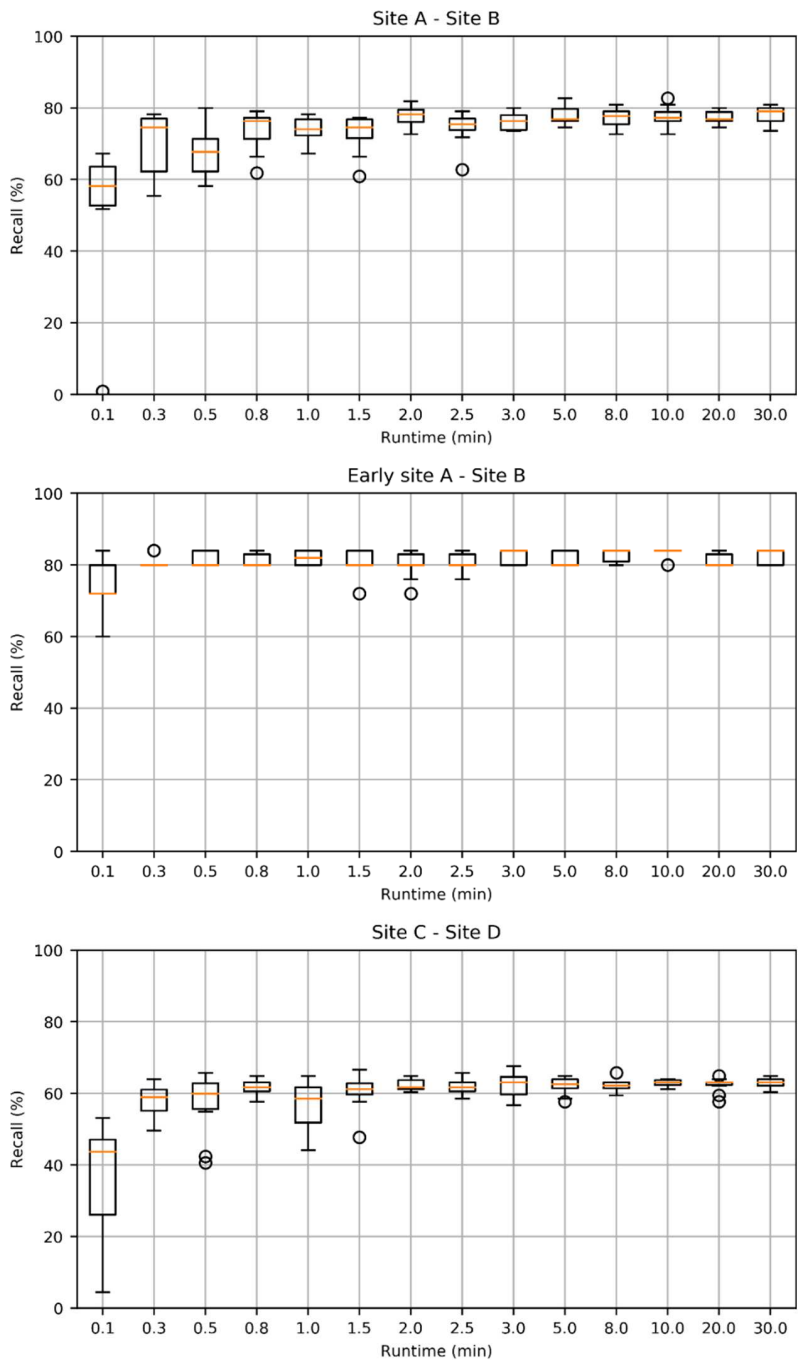

*Figure 18 Recall for different runtimes of the SANA graph-matching algorithm. Tested with simplification algorithm 2 and node similarity function 4 (see section 5.1)*

For SANA, the impact of changing the default weights for WEC and node similarity was investigated (Figure 19). The weights were changed at increments of 0.1 so that the sum of the weights was always 1. In all cases, setting the WEC and node similarity weights close to each other seems to be a reliable strategy for our use case.

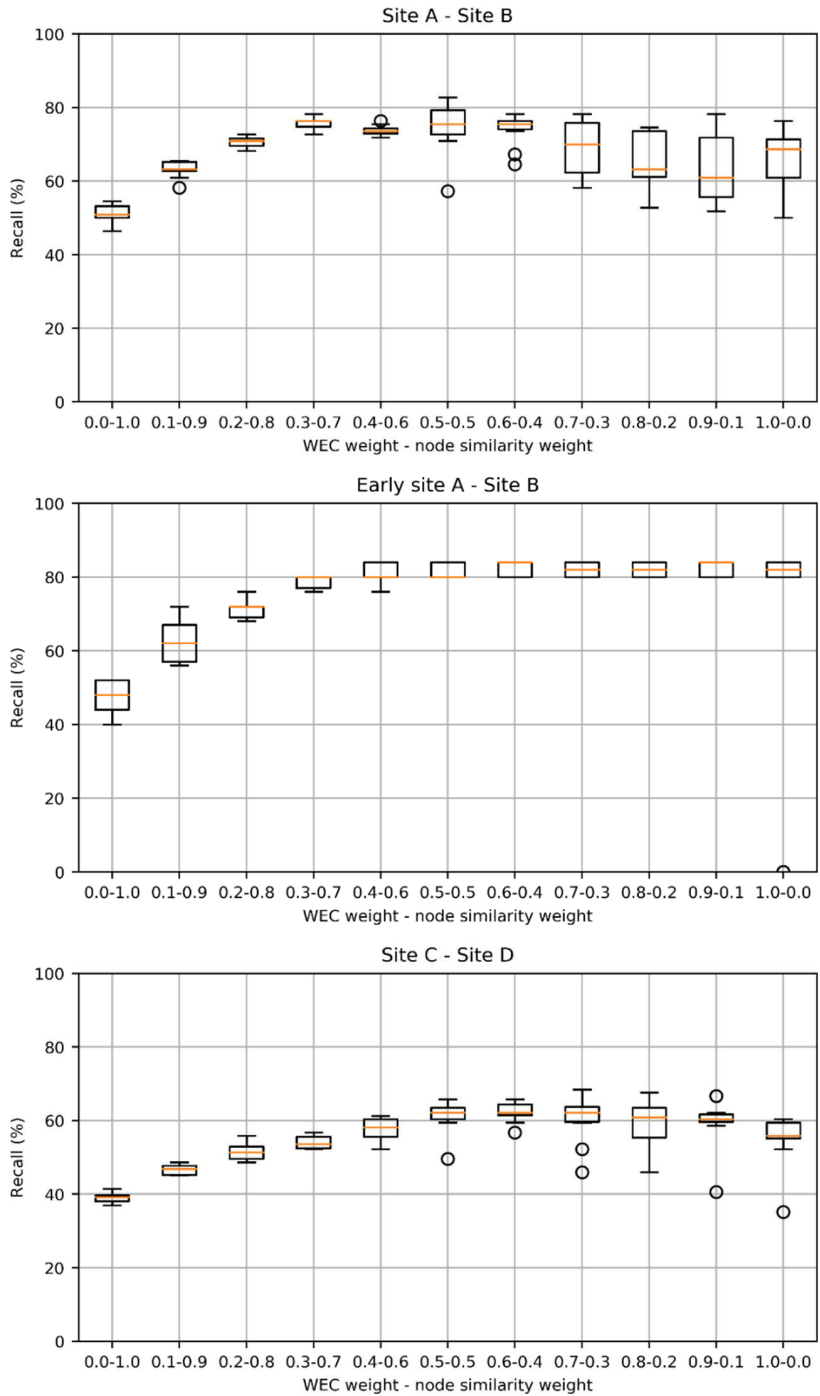

*Figure 19 Empirical evaluation of SANA objective function weights with simplification algorithm 2 and node similarity function 4 (see section 5.1).*

The effect of different node similarity functions depended on the matched sites. In the site C - site D matching, while the best results are still achieved with the more complex similarity function 4, the difference between the similarity functions is not as large as when sites A and B were matched. This could be because sites C and D have more structurally unique areas than sites A and B. Most subsections are included only once or in some cases twice. On the other hand, in the multi-ply board process sites A and B, most areas are repeated three times. As the topological information can guide the matching better with sites C and D, the additional neighborhood and anchor information is not as important, but can still be used to improve the results.

When the completed sites were matched, the best performing algorithm was generally SANA followed by GEDEVO. The recalls with DSPFP and similarity flooding algorithms are considerably lower in comparison. However, when early site A and complete site B were matched, the difference between the algorithms is not as large. The difference in results could be explained by how robust the graph matching algorithms are towards outlier nodes. When sites A and B and sites C and D were matched, the number of outliers was high in both cases. However, there were very few outliers when early site A and site B were matched. With simplification algorithm 2, 25 of the 32 nodes in the early site A graph had a match in the site B graph. This could suggest that metaheuristic algorithms like GEDEVO and SANA are more robust towards a large number of outliers.

## 8. Discussion

In this section, limitations and generality of the proposed methodology is discussed, along with further work to address these issues.

It is notable that the simplification algorithms remove pipelines as well as components along the pipelines, especially valves and also instrumentation in case the source information included instrumentation. The simplification improves the scalability of the algorithms to larger plants, which is of considerable importance, since many matching algorithms can handle only graphs with hundreds of nodes, whereas real industrial plant designs can be much larger. Additionally, the results show improvements in recall with a well-chosen simplification algorithm. However, the components removed by the simplification algorithms are not matched. Nevertheless, if the main process equipment that are the endpoints of a pipeline are properly matched, it is straightforward to match the components along these pipelines. However, the matching is not trivial, since the pipelines may have, for example, a different number of valves. Due to the said benefits of using the simplification algorithms, it is concluded that a two phase approach is worth more detailed investigation in further work. The methodology in this article addresses the first phase of matching simplified overall designs, while further work could address a second phase of matching local details.

There are several process system engineering (PSE) methodologies. As stated in section 2.2, articles on PSE generally lack detailed descriptions of the engineering workflow from the perspective of what design data is created or processed at which phases and how the phases are sequenced. The use case presented in section 1 was obtained from the industrial partner, which is a global leader in the field. Their PSE methodology involves the following distinct, consecutive phases. Firstly, the process structure is defined. After this, the initial parameter estimation is performed for the key process components. It is notable that these steps are not performed for the entire process plant but for one process area at a time; in other words overall optimization of an industrial plant is not attempted but rather the problem is decomposed in a way that has proven practical over the course of numerous delivery projects. Our proposed methodology is strictly limited to the latter phase of initial parameter estimation. Since the initial parameter estimation is a bottleneck task with great impact on the schedule of the plant delivery process, it is proposed that our methodology can offer significant benefits for companies following a PSE methodology similar to the one outlined above. Regarding the findings in section 6, it is concluded that out of the tested algorithms the SANA algorithm is generally expected to give superior performance for the use case in the initial parameter estimation phase in the domains that were investigated, namely multi-board and recycled fiber plants. Further research on additional case studies in these domains could strengthen these conclusions. Repeating this methodology on case studies from other domains would be needed to confirm is SANA or some or algorithm would give the best performance. In general, researchers should determine the best algorithm for an application domain, after which a practitioner can concentrate on preparing the source information required by that algorithm. The source information is company specific, since the sources are the current project and one previous project in the design repository. Thus, practitioners are not expected to evaluate algorithms.

The methodology presented in this article involved the following manual steps. Firstly, data cleaning as illustrated in Figure 3 was performed manually. Automating this, for example with artificial intelligence techniques, is one topic of further research. It is notable that as stated in section 3.1, only a partial data cleaning was performed and reasonably good performance was nevertheless obtained in the results. Secondly, the anchor similarity measure requires the user to set markers, and automating this task may prove difficult. If this is considered laborious by practitioners, the performance of the experiments that did not use the anchor similarity measure may be considered sufficient. Thirdly, manual work to handpick matches was used for evaluation purposes as has been described related to the recall metric and the filter limit – this step is only needed for research to select the best performing algorithm.

The performance of the algorithms in section 6 was expressed in terms of recall of correct matches between components in the sites that comprised our case study. Thus, the numerical results are most appropriate for comparing the different algorithms that were evaluated. Different performance may be encountered by practitioners applying the method to other cases. The complexity of the designs and the similarity of the sites being compared is expected to result in different performance. The methodology is applied at the initial parameter estimation phase, in which a human expert selects a similar previous project, after which the proposed methodology is used to identify similar process components between the current and previous project. It should be emphasized that the recall metric is only used during research to identify the best performing algorithm for a certain domain, such as multi-board plants. In an industrial practice context, that algorithm will be used, so the recall metric will no longer be used. The algorithm will identify a similar process component from a previous project. Currently, it is expected that the expert will reuse the parameter values from that component at his or her discretion. Further research may seek to further automate the reuse and to develop additional metrics for that purpose.

The matching algorithms can cope with the situation, in which the plants contain only some common subsections, in particular when the first graph is only one subsection and the second graph is a complete plant. These cases were tested during the matching of sites A and B and sites early site A and site B. Sites A and B only had similar main processes and contained additional large subsections not found in the other plant. The graph for early site A was made from data available at the initial stages of the project and thus only contained a small subsection of the main process. In general, any approach that tries to segment the problem should consider what kind of information is available in a real industrial practice setting. Our approach doesn't utilize complete information about which equipment belongs to which subsection, because complete sectioning that would be consistent between different projects is not available and manual sectioning would considerably increase manual work. If such information existed, it could be added to equipment as attributes, which would then be compared for similarity. In the absence of such information, defining anchor pairs for equipment that are on the subsection boundaries would be a way to provide the algorithm with partial sectioning information.

## 9. Conclusion

The research goal stated in section 1 was to develop a method to find a similar process components from a previous design project applying graph matching, so that the parameters of the said component could be used for initial parameter estimation of a component in the current project. The developed method consists in the application of several graph transformation and graph matching algorithms as described in section 4. It was validated on an industrial use case. The method is significantly different from previous applications of graph matching technology.

Section 1 elaborated the research goal with four challenges that are specific to this use case. These were addressed as follows:

1. The differences in the matched sites were analyzed with inexact graph matching algorithms and node similarity measures proposed in section 4.2, which allowed matches between similar nodes even if they were not an exact match. An empirical evaluation of the impact of exploiting these similarity measures was presented in section 6. Additionally, a solution for fixing a subset of the possible errors and inconsistencies in source data was presented in section 3.1 Figure 3.
2. The comparison between the early site A and the detailed site B (Figure 15) demonstrated the applicability of the methodology in the beginning of the project when only early phase design information is available
3. The graph simplification techniques in section 4.1 significantly reduced the size of the graph and thus the computational complexity as shown in Table 11. The results in section 6 show that the best performance was often obtained by the most aggressive simplification, involving removal of all pipes.
4. In section 4.2, edge labels and direction were incorporated to the node similarity scores as follows. The edge label and direction were considered when calculating the neighbor score. Only the edge direction was considered in calculating the anchor score.

Acknowledgement

Authors acknowledge funding support from Aalto University and Semantum Oy.